\documentclass{article} 
\usepackage{iclr2026_conference,times}
\iclrfinalcopy 

\usepackage{amsmath,amsfonts,bm}

\def\eqref#1{equation~\ref{#1}}

\def\1{\bm{1}}

\DeclareMathAlphabet{\mathsfit}{\encodingdefault}{\sfdefault}{m}{sl}
\SetMathAlphabet{\mathsfit}{bold}{\encodingdefault}{\sfdefault}{bx}{n}

\usepackage{hyperref}
\usepackage{url}
\usepackage{amssymb} 
\usepackage{amsmath}
\usepackage{multirow}
\usepackage{graphicx}
\usepackage{booktabs}
\usepackage{subcaption}
\usepackage{soul}
\usepackage[most]{tcolorbox}   
\usepackage{enumitem}          
\usepackage[T1]{fontenc}
\usepackage[utf8]{inputenc}
\usepackage[most]{tcolorbox}
\usepackage{caption}
\usepackage{wrapfig} 
\usepackage{adjustbox}
\usepackage{subcaption}
\usepackage{caption}

\usepackage[ruled,noend]{algorithm2e}
\DontPrintSemicolon
\SetAlgoLined
\SetKwInput{KwIn}{Input}
\SetKwInput{KwOut}{Output}
\SetKwInput{KwData}{Symbols}
\SetKwInput{KwParams}{Hyperparams}
\SetKwInput{KwRet}{Retained Intermediates}

\newcommand{\Stage}[1]{\par\smallskip\noindent\textbf{#1}\ \ignorespaces}
\usepackage{tikz}
\usetikzlibrary{arrows.meta,positioning,fit,shapes.misc}
\usepackage{soul}
\usepackage{xcolor}
\definecolor{lightblue}{RGB}{210,230,255}
\sethlcolor{lightblue}   

\title{ChainMPQ: Interleaved Text-Image Reasoning Chains for Mitigating Relation Hallucinations}

\author{Yike Wu$^{1}$\quad\quad 
Yiwei Wang$^{2}$\thanks{Corresponding author.}\quad\quad
Yujun Cai$^{1, 3}$\\
$^{1}$University of Queensland\quad\quad 
$^{2}$University of California, Merced\quad\quad 
$^{3}$Ant Group\\
\texttt{yikewu66@gmail.com};\  \ \ \texttt{yiweiwang2@ucmerced.edu}\\
\href{https://yike-wu.github.io/chainmpq-project/}{\textcolor{magenta}{\texttt{https://yike-wu.github.io/chainmpq-project}}}
}
\begin{document}

\maketitle

\begin{abstract}
While Large Vision-Language Models (LVLMs) achieve strong performance in multimodal tasks, hallucinations continue to affect their reliability. Among the three categories of hallucinations, which include object, attribute, and relation, relation hallucinations account for the largest proportion but have received the least  attention. To address this challenge, we propose \textbf{ChainMPQ} (\textbf{M}ulti-\textbf{P}erspective \textbf{Q}uestions guided Interleaved Text-image Reasoning \textbf{Chain}), a training-free method that improves relational inference in LVLMs by utilizing accumulated textual and visual memories. ChainMPQ first extracts subject and object keywords from the question to enhance the corresponding image regions. It then constructs multi-perspective questions that focus on the three core components of a relationship: the subject, the object, and the relation that links them. These questions are sequentially input to  the model, with textual and visual memories from earlier steps providing supporting context for subsequent ones, thereby forming an interleaved chain of image and text that guides progressive relational reasoning. Experiments on multiple LVLMs and benchmarks show that ChainMPQ substantially reduces relation hallucinations, while ablation studies further validate the effectiveness of its three core modules.

\end{abstract}

\vspace{-10pt}
\section{Introduction}

\begin{figure}[t]
    \centering
    \includegraphics[width=1.0\textwidth]{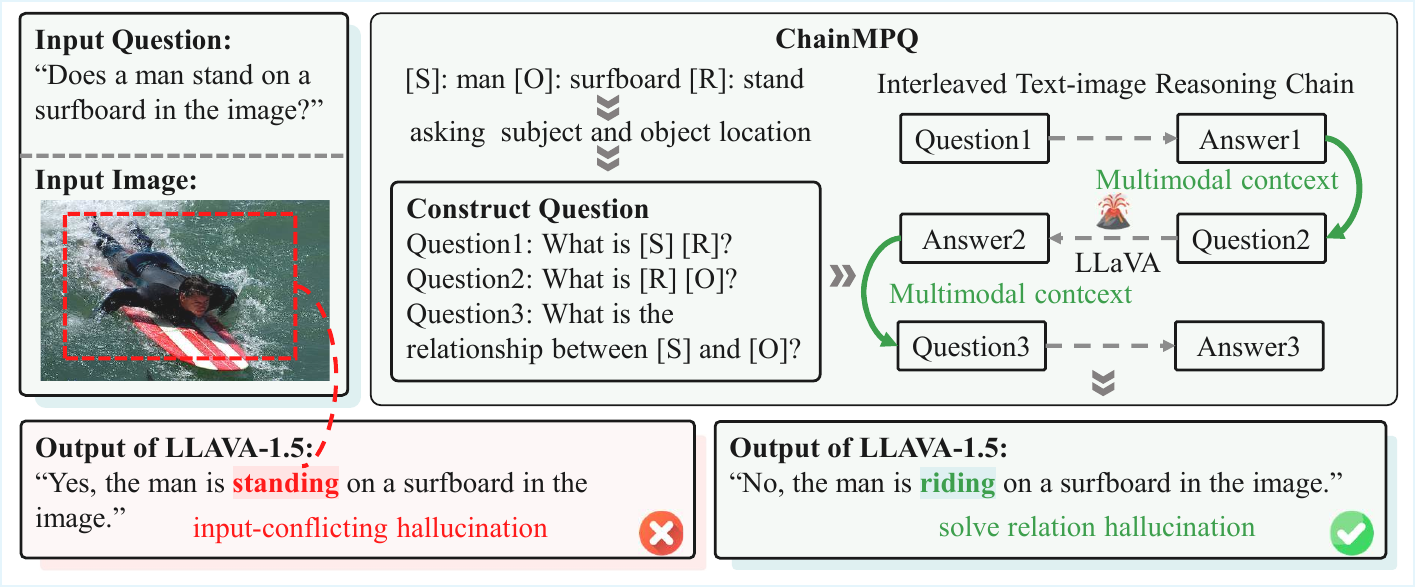}
    \caption{This figure illustrates a case in which an LVLM exhibits a relation hallucination, misrecognizing the true relation "riding" as "standing". With ChainMPQ, the model is guided to reason step by step and infer the correct relation.}
    \label{introduction}
    \vspace{-16pt}
\end{figure}

Large Language Models (LLMs) have made substantial advances in language understanding, generation, and reasoning~\cite{touvron2023llama2openfoundation,touvron2023llamaopenefficientfoundation,bai2023qwentechnicalreport}. Extending these capabilities with visual inputs, Large Vision-Language Models (LVLMs) achieve strong performance on a range of multimodal tasks, including image captioning and visual question answering~\cite{ye2024mplugowlmodularizationempowerslarge,li2023mimicitmultimodalincontextinstruction,bai2023qwenvlversatilevisionlanguagemodel,li2023blip2bootstrappinglanguageimagepretraining,dai2023instructblip,liu2024improved}. Nevertheless, LVLMs still exhibit hallucinations, producing outputs that contradict or overlook the visual evidence. Hallucinations in LVLMs can be broadly categorized into object, attribute, and relation hallucinations~\cite{bai2024hallucination}. Object hallucinations refer to failures in recognizing entities, while attribute hallucinations involve misidentifying properties such as color or shape. Relation hallucination occurs when models correctly recognize objects but fail to infer the relationship between them, as shown in Figure~\ref{introduction}.

 Although prior work has substantially reduced object and attribute hallucinations through preference optimization~\cite{wu2025mitigating,yang2025mitigating}, contrastive decoding~\cite{wu2025generate,zhang2025self}, and intermediate-layer modification~\cite{jiang2024interpreting,jiang2025devils,wang2024mllm}, relation hallucination, which accounts for nearly 40\% of all hallucinations~\cite{zheng2024reefknot}, remains a salient challenge.  Existing efforts to mitigate relation hallucinations have made some progress: dataset-driven approaches~\cite{xie2025fg} focus on constructing high-quality training and fine-tuning data, prompt-engineering methods~\cite{wu2025mitigating} employ constrained perception prompts, Detect-then-Calibrate~\cite{zheng2024reefknot}  selectively adjust the final logits with intermediate layer, and Triplet  Description~\cite{wu2025unifiedtripletlevelhallucinationevaluation} converts the image into triplets, allowing the model to answer questions based on these structured descriptions instead of the raw image.

Despite their promising results, existing methods targted at relation hallucinations treat relational reasoning as single-step inference, expecting models to simultaneously identify entities and determine relationships. This approach is likely to produce errors because it relies heavily on language priors rather than systematic visual analysis. Human relational reasoning, by contrast, follows a more structured process: first locating and identifying relevant objects, then examining their interactions, and finally synthesizing visual evidence to draw conclusions about relationships. This multi-step approach allows for more careful consideration of visual information and reduces the likelihood of inference errors.

Inspired by human reasoning process and drawing insights from Interleaved Modal Chain-of-Thought (ICoT)~\cite{gao2025interleaved}, we propose \textbf{ChainMPQ} (\textbf{M}ulti-\textbf{P}erspective \textbf{Q}uestions guided Interleaved Text-image Reasoning \textbf{Chain}), a training-free framework that decomposes relational inference into manageable steps while maintaining relevant reasoning through accumulated multimodal memory. Our approach addresses the core challenge of relation hallucination by combining systematic question decomposition with progressive multimodal reasoning.

Specifically, ChainMPQ first extracts subject and object keywords from the question and uses them to drive cross-attention. Next, relations are decomposed into three components, and multi-perspective questions are constructed by masking one component at a time. Finally, the interleaved chain is built by sequentially feeding the constructed questions into the model, using previous answers as textual context and earlier relevant visual tokens to adjust later attention maps. This process enables the model to reason over prior textual and visual memories and progressively resolve the relation. We demonstrate the effectiveness of ChainMPQ on four advanced and widely used models, where it consistently reduces relation hallucinations on relation-focused benchmarks. Our contributions are summarized as follows:
\begin{enumerate}
    \item We introduce a subject–object–relation decomposition to generate multi-perspective questions, encouraging the model to focus on each core element of a relationship.
    \item We design an interleaved chain mechanism that transfers textual and visual memories by using answers and attention maps from earlier steps to refine subsequent reasoning, thereby enabling progressive relational inference.
    \item We apply ChainMPQ to multiple LVLMs and relation-focused benchmarks, demonstrating consistent reductions in relation hallucinations.
\end{enumerate}

\section{Related Work}
\subsection{Large Vision-Language Models (LVLMs)}
Large Vision-Language Models (LVLMs) enhance the capabilities of traditional Large Language Models (LLMs) by incorporating visual inputs, thereby enabling them to perform complex multimodal tasks such as image captioning and visual question answering. Prominent LVLMs, including InstructBLIP~\cite{dai2023instructblip} and LLaVA~\cite{liu2024improved}, integrate pre-trained vision encoders with language models, effectively bridging the gap between the visual and textual modalities.

While these models have achieved remarkable progress in visual-language understanding~\cite{lai2024lisareasoningsegmentationlarge,laurençon2024buildingbetterunderstandingvisionlanguage,li2025enhancingvisionlanguagecompositionalunderstanding,peng2024synthesizediagnoseoptimizefinegrained}, they continue to face challenges of hallucinations, particularly in the recognition of relationships~\cite{zheng2024reefknot,nie2024mmrel} between objects. Our work extends these models by specifically addressing the problem of relation hallucinations, employing a step-by-step reasoning process guided by both textual and visual information.

\subsection{Relation Hallucination in LVLMs}
Despite significant advancements in mitigating object and attribute hallucinations, relation hallucination remains a critical and underexplored issue~\cite{liu2024surveyhallucinationlargevisionlanguage,wu2024evaluating}. These hallucinations occur when a model accurately detects objects but fails to correctly identify the relationships between them, leading to misleading or incorrect responses in tasks such as Visual Question Answering (VQA)~\cite{shahgir2024illusionvqachallengingopticalillusion}. The significance of addressing relation hallucinations is highlighted by their prevalence and studies indicate that nearly 40\% of hallucinations in LVLMs are related to relations between objects~\cite{zheng2024reefknot}, yet this issue has not received as much attention as object or attribute hallucinations.

Recent research has proposed several evaluation benchmarks aimed at assessing the handling of inter-object relations. For example, MMRel~\cite{nie2024mmrel} features over 10k question-answer pairs across multiple domains, specifically designed to evaluate spatial, action, and comparative relations. Additionally, R-Bench~\cite{wu2024evaluating} includes a diverse set of questions that balance perception and cognition tasks, exposing  limitations in the relation reasoning capabilities of contemporary models. Tri-HE~\cite{wu2025unifiedtripletlevelhallucinationevaluation} further extends evaluation by using triplets to represent knowledge, which allows it to assess both object and relation hallucinations at the same time. These benchmarks are valuable for systematically evaluating and enhancing the management of relation hallucinations. For considerations of dataset size and usage frequency, we evaluate our method using MMRel and R-Bench, which are larger and more widely used.

\subsection{Interleaved Modal Chain-of-Thought (ICoT)}

Interleaved Modal Chain-of-Thought (ICoT)~\cite{gao2025interleaved} extends the Chain-of-Thought (CoT) reasoning paradigm~\cite{wei2022chain} to multimodal tasks by  integrating visual and textual reasoning steps. While traditional CoT methods, which primarily focus on text, struggle to handle dynamic visual state transitions, ICoT addresses this limitation by progressively updating intermediate visual states throughout the reasoning process. This enables the model to maintain coherent and grounded reasoning across both modalities. The dynamic interaction between visual and textual information in ICoT closely mirrors human cognitive processes, where visual and textual elements evolve concurrently to support decision-making. Recent advances in multimodal reasoning, including approaches such as Uni-CoT~\cite{qin2025uni} and visual reasoning frameworks~\cite{lin2025mind}, emphasizes the importance of integrating visual feedback into reasoning. ICoT offers a flexible, scalable, and efficient framework for multimodal reasoning, establishing a new basic method for future research in this field. Given the promising potential of ICoT, we leverage its principles to build our own methods.

\section{Methods}

\subsection{PROBLEM FORMULATION}
We address relation hallucination in visual question answering, where models correctly identify entities but fail to infer their relationships. Given an image $I$ and a relational question $Q$, the task is to generate an accurate Yes/No answer $A$. Relation hallucination occurs when a model detects both subject and object entities but provides incorrect relational judgments. 

\subsection{Overview}


\begin{figure}[t]
    \vspace*{-16pt}
    \centering
    \includegraphics[width=1.0\textwidth]{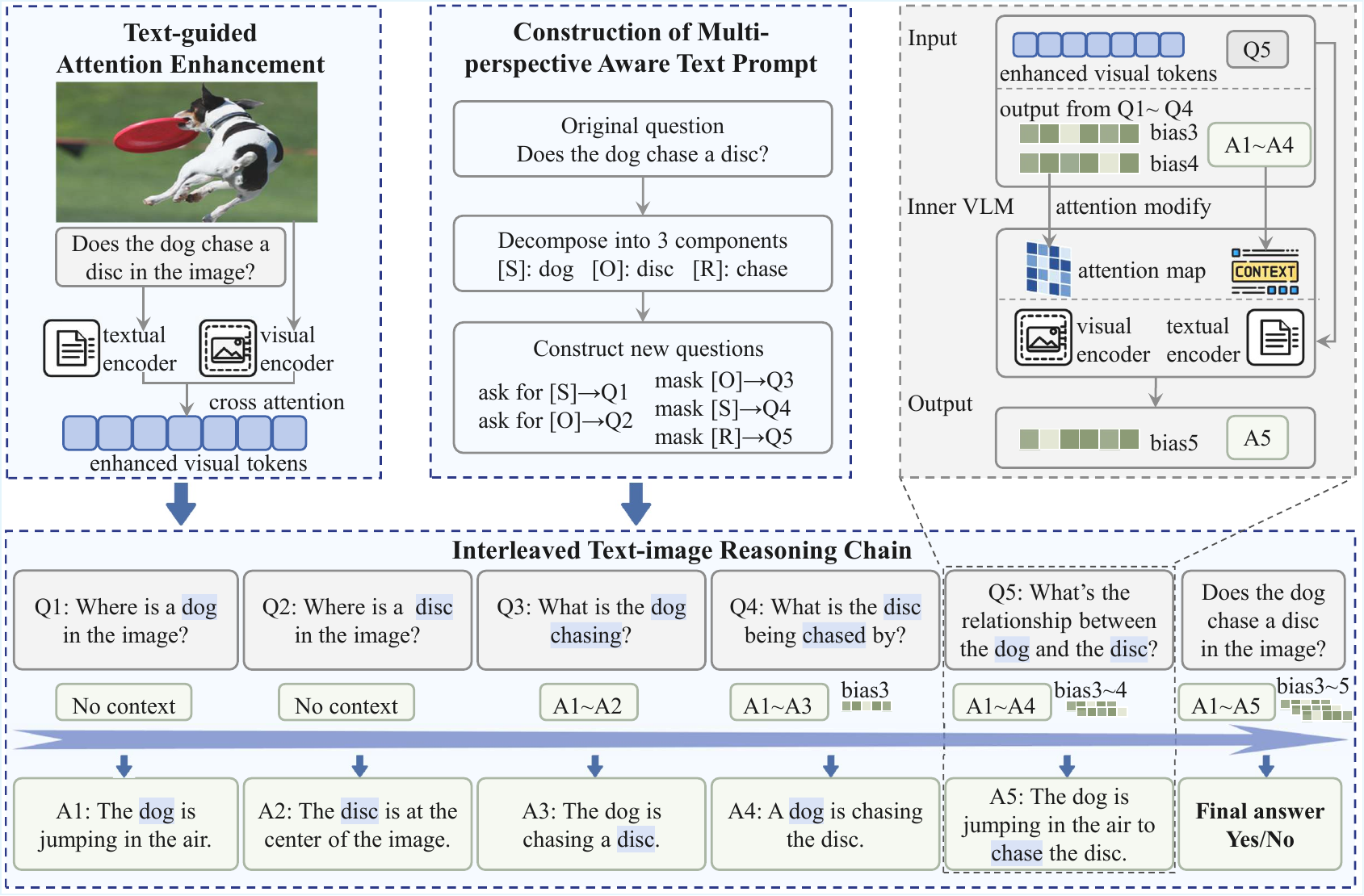}
    \caption{Overview of our proposed ChainMPQ. It comprises three modules. Text-guided Attention Enhancement: extracts subject, object, and relation, using cross-attention to emphasize relation-relevant visual regions; Multi-Perspective Aware Text Prompt: constructs five new questions based on these elements from different perspectives; Interleaved Text-image Reasoning Chain: sequentially inputs the questions, using each answer $A_i$ and its top-K active visual tokens to form mask $M_i$ as a bias when calculating subsequent attention maps. The original question is then answered to produce the final output and evaluation metrics.}
    \label{framework}
    \vspace{-8pt}
\end{figure}

ChainMPQ is a training-free framework designed to mitigate relation hallucinations through progressive multimodal reasoning. Unlike existing approaches that rely solely on textual prompting, our method leverages both textual answers and visual attention patterns across reasoning steps. For procedural details of the algorithm, see Algorithm~\ref{alg:chainmpq_slim}.

The framework operates in three stages: (1) enhancing visual representations of entities mentioned in the question through text-guided attention mechanisms, (2) decomposing the original relational query into multiple complementary questions that target individual components of the relationship, and (3) constructing an interleaved chain where accumulated textual and visual information guides subsequent reasoning. This progressive approach enables the model to systematically analyze relationships rather than making immediate judgments based on surface patterns.

\subsection{Text-guided Attention Enhancement}
Accurate relational reasoning requires precise entity localization. We extract subject and object keywords from the input question using the spaCy NLP toolkit. These keywords are encoded to obtain representations $X \in \mathbb{R}^{N \times d_t}$, where $N$ denotes the number of keywords (typically two, corresponding to the number of subject and object), and $d_t$ represents the text feature dimension.

The input image is processed by the visual encoder to obtain visual features $V \in \mathbb{R}^{M \times d_v}$, where $M$ is the number of image patches and $d_v$ is the dimension of the visual features. We apply cross-attention to enhance image regions corresponding to the extracted keywords, where the image features $V$ serves as the Query vector and the keyword text $X$ as both the Key and Value vectors:
\begin{equation}
V' = \operatorname{softmax}\left(\frac{V X^\mathrm{T}}{\sqrt{d_t}}\right) X
\end{equation}

This operation produces enhanced visual tokens $V'$ that emphasize subject and object regions, establishing a foundation for accurate relational inference in subsequent steps.

\subsection{Construction of Multi-Perspective Aware Text Prompt}
We decompose the original question into five complementary questions that examine different aspects of the relationship. Specifically, the original input question typically contains three components: the subject, the object, and the relation between them. The first two questions focus on entity localization by asking about the location of the subject and object respectively. The remaining three questions employ a masking strategy where one component is masked while the other two are used to construct new queries.

To be specific, we mask the object to generate a question about what the subject is interacting with, mask the subject to ask what the object is being affected by, and mask the relation to inquire about the general relationship between the entities. For example, "Does the dog chase a disc in the image?" becomes five questions: entity localization ("Where is the dog/disc?") and three relation-focused queries generated through systematic masking, as shown in Figure~\ref{framework}.

This decomposition strategy encourages the model to analyze individual components before making final relational judgments, reducing reliance on language priors and improving systematic reasoning.

\subsection{Interleaved Text-Image Reasoning Chain}
\begin{algorithm}[t]
\small
\caption{ChainMPQ}
\label{alg:chainmpq_slim}

\KwIn{Image $I$, relational question $Q$}
\KwOut{Final answer $A$}
\KwParams{$\lambda$, $k_{\max}$}
\KwData{Textual context $\mathcal{T}$, visual memory $\mathcal{V}$}

\textbf{Module 1: Enhance visual tokens}\\
\Indp
Extract keywords $K$ from $Q$;\;
Encode $V \leftarrow \mathrm{VisualEncoder}(I)$ and 
$X \leftarrow \mathrm{TextEncoder}(K)$;\;
Obtain enhanced visual tokens $V' \leftarrow \mathrm{CrossAttn}(X, V)$.\;
\Indm

\textbf{Module 2: Construct sub-questions}\\
\Indp
Decompose $Q$ into $[S], [O], [R]$;\;
Generate complementary sub-questions $Q_{1..5}$.\;
\Indm

\textbf{Module 3: Interleaved text–image reasoning chain}\\
\Indp
Q1 and Q2: localize $[S]$ and $[O]$ using $V'$, update $\mathcal{T}$;\;
Q3 to Q5: ask relation-focused questions and build top-$k$ mask $M_i$ with $k \le k_{\max}$;\;
Apply bias scaled by $\lambda$, update $\mathcal{T}$ and $\mathcal{V}$;\;
Answer original question: decode with $V'$, $\mathcal{T}$ and $\mathcal{V}$ to produce $A$.\;
\Indm

\end{algorithm}

Unlike previous text-only prompting methods~\cite{wu2025mitigating}, we sequentially process the enhanced
visual tokens from Section 3.3 and constructed questions from Section 3.4 using an interleaved chain that transfers both textual and visual information across reasoning steps. For each question $Q_i$, we encode it using the text encoder to obtain question tokens, while maintaining a context $C_i$ that accumulates answers from previous steps.



The model generates answer $A_i$ for question $Q_i$ using the enhanced visual tokens $V'$, current context $C_i$, and any accumulated visual memory from earlier steps. The first and second questions are answered directly without adding any contextual information. Starting from the third question, we extract attention weights associated with the keyword tokens in each question from the final $n$ decoder layers to capture the model's focus on visual regions: 
\begin{equation}
\label{attention}
\text{Attn}_i = \frac{1}{|T| \cdot n} \sum_{t \in T} \sum_{\ell = L - n}^{L - 1} \text{Attn}^{(\ell)} [t, :]
\end{equation}
where T contains indices of keyword tokens, L is the total number of decoder layers, and $\text{Attn}^{(\ell)}$ represents the attention matrix at layer $\ell$.

We select the top-k visual tokens with highest attention scores using an entropy-based adaptive strategy that adjusts $k$ based on attention concentration:
\begin{equation}
\label{topk}
k = \left| k_{\text{max}} \cdot \hat{H}(\text{Attn}_i) \right|
\qquad
I_{\text{topk},i} = \text{TopK}(\text{Attn}_i, k)
\end{equation}
where $\hat{H}(\text{Attn}_i)$ represents the normalized entropy of the attention distribution, $I_{\text{topk},i}$ denotes the set of visual tokens most attended to for the $i$-th question and $k_{max}$ is set to 20 in our experiments.

The selected tokens form a bias mask $M_i$ that guides attention in subsequent steps:


\begin{equation}
\label{m}
M_{i,j}= 
\begin{cases}
    \frac{\text{Attn}_{i,j}}{\sum_{j \in I_{\text{topk},i}} \text{Attn}_{i,j}}  & j\in I_{\text{topk},i}  \\
    \text{0} & \operatorname{other cases}
\end{cases}
\end{equation}
where entries corresponding to $I_{\text{topk},i}$ are assigned normalized attention scores and all others are set to zero, and $M_{i,j}$ denoting the $j$-th element in $M_i$.

For subsequent questions, the attention mechanism incorporates this visual bias:
\begin{equation}
\label{modify attention}
\alpha_i = \lambda\cdot \mathit{Conf}_{\text{prev}_i} \qquad \text{Attn}_{i+1} = \operatorname{softmax}\!\left(\frac{QK^{\top}}{\sqrt{d_k}} + \alpha_i\cdot M_i\right) V 
\end{equation}
where $\alpha_i$ is a confidence-based weight that increases with answer certainty.

For multi-round previous visual bias, we  calculate a weighted average:
\begin{equation}
\label{attention-fuse}
\text{Attn}_{i+1} = \operatorname{softmax}\!\left(\frac{QK^{\top}}{\sqrt{d_k}} + \frac{1}{\sum\limits_{j=3}^{i} \alpha_j} \sum_{j=3}^{i} \alpha_j \cdot M_j\right) V
\end{equation}
This mechanism enables the model to maintain visual focus across reasoning steps while progressively building understanding of the relationship. The accumulated multimodal evidence culminates in a final answer that reflects systematic relational analysis rather than surface-level pattern matching.



\section{Experiments}
\subsection{Experimental Setup}
\textbf{Evaluated LVLMs.} We evaluate our method on four open-source LVLMs: LLaVA-1.5-7B~\cite{liu2024improved}, InstructBLIP-7B~\cite{dai2023instructblip}, Qwen2.5-VL-7B~\cite{bai2025qwen25vltechnicalreport}, and InternVL3.5-8B~\cite{wang2025internvl35advancingopensourcemultimodal}. The first two models are widely used and represent classic LVLM designs. The last two models are more recent and show stronger performance in many multimodal tasks. LLaVA-1.5 combines a CLIP visual encoder with Vicuna-7B for multimodal reasoning, while InstructBLIP-7B follows the BLIP-2 design, linking a frozen vision encoder to a language model through a Q-Former. Qwen2.5-VL is built on the Qwen2.5 language model and uses a redesigned ViT architecture to provide strong visual understanding.
InternVL3.5 integrates an InternViT visual encoder with a Qwen3 series language model to support robust image reasoning.
These distinct architectures allow us to assess the generalization of our method across different design paradigms.

\noindent

\textbf{Dataset and Benchmarks.} 
Commonly used benchmarks for visual question answering, such as POPE~\cite{li2023evaluating} and CHAIR~\cite{rohrbach2018object}, primarily focus on object hallucinations. Since our task aims to alleviate relation hallucinations, we evaluate on two widely used benchmarks specifically designed for this purpose: MMRel~\cite{nie2024mmrel} and R-Bench (image-level)~\cite{wu2024evaluating}. The  datasets details are illustrated in Appendix. 
\begin{table*}[t!]
\centering
\caption{Results on MMRel and R-Bench benchmark. Higher (↑) accuracy, precision, and F1 indicate better performance. The best results are bolded, and the second-best are underlined.}
\label{main-ex}
\begin{tabular}{l l c c c c c c}
\toprule
\multirow{2}{*}{Model} & \multirow{2}{*}{Method} & \multicolumn{3}{c}{MMRel} & \multicolumn{3}{c}{R-Bench} \\
\cmidrule(lr){3-5} \cmidrule(lr){6-8}
                      &                        & Acc & Prec & F1 & Acc & Prec & F1 \\
\midrule
\multirow{5}{*}{LLaVA-1.5} 
                      & Vanilla                & 59.02 & 56.81 & 66.40 & 71.23 & 64.27 & 77.28 \\
                      & Prompting~\cite{wu2025mitigating} & 62.33 & 60.58 & 68.07 & \underline{75.86} & 67.28 & \underline{79.60} \\
                      & Calibrate~\cite{zheng2024reefknot} & \underline{63.50} & \underline{60.79} & \underline{70.54} & 74.28 & \underline{67.86} & 78.31 \\
                      & CoT ~\cite{wei2023chainofthoughtpromptingelicitsreasoning}     &  61.88 & 59.14 & 66.25 & 72.91 & 65.88 & 77.87 \\
                      & Ours                   & \textbf{65.20} &  \textbf{64.75} & \textbf{71.21} &
                      \textbf{76.04} & \textbf{72.03}& \textbf{81.54} \\
\midrule
\multirow{5}{*}{InstructBLIP}
                      & Vanilla                & 57.58 & 55.32 & 66.79 & 69.31 & 62.76 & 76.04 \\
                      & Prompting~\cite{wu2025mitigating} & \underline{64.52} & \underline{62.28} & \underline{71.23} & \underline{73.65} & \underline{68.74} & \underline{80.21} \\
                      & Calibrate~\cite{zheng2024reefknot}  & 62.05 & 61.02 & 69.85 & 72.96 & 66.23 & 78.54 \\
                      & CoT ~\cite{wei2023chainofthoughtpromptingelicitsreasoning}      & 59.90 & 57.12 & 65.41 & 72.15 &65.79 & 78.45 \\
                      & Ours & \textbf{65.14} & \textbf{64.12} & \textbf{74.12} & \textbf{75.86} & \textbf{70.59} & \textbf{81.12} \\
\midrule
\multirow{4}{*}{Qwen2.5-VL}
                      & Vanilla                & 66.10 & 63.07 & 72.14 & 79.85 & 75.19 & 80.88 \\
                      & Prompting~\cite{wu2025mitigating} & 70.01 & \underline{68.59} & 75.23 & \underline{81.02} & \underline{78.38} & \underline{82.46} \\
                      & Calibrate~\cite{zheng2024reefknot} & \underline{71.36} & 67.20 & \underline{76.28} & 80.98 & 77.05 & 81.52 \\
                      & CoT ~\cite{wei2023chainofthoughtpromptingelicitsreasoning}     & 67.35 & 65.49 & 72.08 & 80.24 & 76.94 & 81.64 \\
                      & Ours                   & \textbf{73.52} & \textbf{69.65} & \textbf{77.45} & \textbf{83.92} & \textbf{80.23} & \textbf{84.63} \\
\midrule
\multirow{4}{*}{InternVL3.5}
                      & Vanilla                & 71.44 & 67.07 & 75.86 & 82.33 & 78.87 & 83.21 \\
                      & Prompting~\cite{wu2025mitigating} & 73.01 & 67.65 & 77.12 & \underline{83.97} & \underline{80.31} &\underline{84.87} \\
                      & Calibrate~\cite{zheng2024reefknot} & \underline{74.45} & \underline{71.46} & \underline{78.64} & 82.65 & 79.82 & 83.75 \\
                      & CoT ~\cite{wei2023chainofthoughtpromptingelicitsreasoning}      & 72.10 & 68.74 & 76.35 & 82.87 & 79.41 & 84.02 \\
                      & Ours                   & \textbf{75.21 }& \textbf{72.54} & \textbf{78.65} & \textbf{85.05} & \textbf{82.85} & \textbf{85.91} \\
\bottomrule
\end{tabular}
\end{table*}

\noindent

\textbf{Baselines.} 
In addition to a standard multimodal large model, we compare our approach with standard Chain-of-Thought prompting~\cite{wei2023chainofthoughtpromptingelicitsreasoning} and several training-free methods specifically designed to mitigate relation hallucinations. These include Constraint-Aware Prompting~\cite{wu2025mitigating}, which relies on text-based prompting, and Detect-then-Calibrate~\cite{zheng2024reefknot}, which calibrates the final output layer using hidden states from intermediate layers. We chose them because they represent two main ideas in this area: text-prompt-based reasoning and output calibration. Comparing with them allows us to show the benefit of our full text-image reasoning chain in a clear way. The performance of these baseline methods is obtained by re-implementing their experiments on our chosen datasets and experimental setup using the latest public code and their official instructions.

\subsection{Main Results}
\textbf{Overall Performance.} Table~\ref{main-ex} shows ChainMPQ consistently outperforms baselines across both models and benchmarks. On MMRel, ChainMPQ achieves 65.20\% accuracy with LLaVA-1.5, improving over the best baseline by 1.7\%. Similar improvements are observed with InstructBLIP (65.14\% vs 64.52\%). On Qwen2.5-VL and InternVL3.5, which already have higher base performance, ChainMPQ also improves accuracy from 71.36\% to 73.52\% and from 74.45\% to 75.21\% respectively. The method shows particularly strong precision gains (4.17\% higher than the best baseline using LLaVA in R-Bench), indicating reduced false positive relation predictions while simultaneous gains in F1 further confirm
that the method improves overall reliability without sacrificing recall.

\textbf{Cross-Model Generalization.} ChainMPQ demonstrates consistent improvements across different architectures (LLaVA, InstructBLIP, Qwen-VL and InternVL), suggesting the approach is model-agnostic rather than exploiting specific architectural features.


\subsection {Performance Experiment}
To further improve the efficiency of ChainMPQ, we explored two optimization methods to achieve a trade-off between accuracy and latency.
\begin{enumerate}
    \item \textbf{Light1:} We keep only Q1, Q2, and Q5. Since Q1 and Q2 don't need to pass text or visual information to each other, they can run in parallel. When answering Q5, we use the answers of Q1 and Q2 as the text context, and their bias masks M1 and M2, constructed from the attention map, as the visual context.
    \item \textbf{Light2:} We keep only Q3, Q4, and Q5, and all other settings remain the same. This means the reasoning chain starts from Q3 with no text or visual context, and the remaining steps (from Q3 to Q5) follow the same procedure described in Section 3.5.
\end{enumerate}
The results in Table~\ref{tab:acc-latency} show that the Light1 has the highest $\Delta \text{Acc} / \Delta \text{Time}$
, having a higher accuracy improvement in a shorter time, which means it achieves the best overall trade-off. Therefore, when accuracy is the top priority, we use the full ChainMPQ. When a balance between accuracy and latency is needed, Light1 is the preferred choice.
\begin{table}[t]
\centering
\caption{Accuracy and latency comparison of different methods on LLaVA-1.5-7B.}
\label{tab:acc-latency}
\begin{tabular}{l l c c c}
\toprule
Benchmark & Method & Acc (\%) & Time (s/sample) & $\Delta$Acc / $\Delta$Time \\
\midrule
\multirow{4}{*}{MMRel}
    & Vanilla       & 59.02 & 0.9 & -- \\
    & Full ChainMPQ & 65.20 & 3.3 & 2.58 \\
    & Light1        & 63.78 & 1.5 & \textbf{7.93} \\
    & Light2        & 64.25 & 2.1 & 4.36 \\
\midrule
\multirow{4}{*}{R-Bench}
    & Vanilla       & 71.23 & 0.9 & -- \\
    & Full ChainMPQ & 76.04 & 3.3 & 2.00 \\
    & Light1        & 74.50 & 1.5 & \textbf{5.45} \\
    & Light2        & 75.08 & 2.1 & 3.21 \\
\bottomrule
\end{tabular}
\end{table}

\subsection{Ablation Study}


We conduct ablation studies to evaluate the impact of key components and hyperparameters in our method, using LLaVA-1.5 on the MMRel benchmark.
\subsubsection{Ablation study of core components.}
We examine the contributions of ChainMPQ’s core components. To assess the role of "Text-guided Attention Enhancement", we remove the enhancement of visual tokens in subject and object regions, such that in the  multimodal chain we use $V$ rather than $V'$.To test the importance of the "Construction of Multi-perspective Aware Text Prompts", we replace the five constructed questions with only the relationship question (Q5), meaning that the model only answers Q5 and then the original question. And The model answers the original question using the text context and visual bias produced when answering Q5. To evaluate the contribution of the "Interleaved Text-image Reasoning Chain", we remove the multimodal chain and only use the text context from the previous answer, which means that we keep the entire multi-turn text context, not only the last answer.

\begin{figure*}[t]
\centering
\begin{subfigure}{0.48\textwidth}
  \centering
  \renewcommand{\arraystretch}{1.2} 
  \begin{tabular}{cccc}
   \toprule
    Model Variants        & Acc   & Prec   & F1    \\
    \midrule
    LLaVA-1.5             & 59.02 & 56.81  & 66.40 \\
    \midrule
    \textbf{ChainMPQ(Full)} & \textbf{65.20} & \textbf{64.75} & \textbf{71.21} \\
    w/o Enhancement       & 64.06 & 63.25  & 69.42 \\
    w/o Multi-perspective & 61.52 & 60.84  & 67.53 \\
    w/o Interleaved       & 62.12 & 61.47  & 68.01 \\
    \bottomrule
  \end{tabular}
  \caption{Ablation study with different model variants.}
  \label{ablation}
\end{subfigure}\hfill
\begin{subfigure}{0.48\textwidth}
  \centering
  \includegraphics[width=\linewidth]{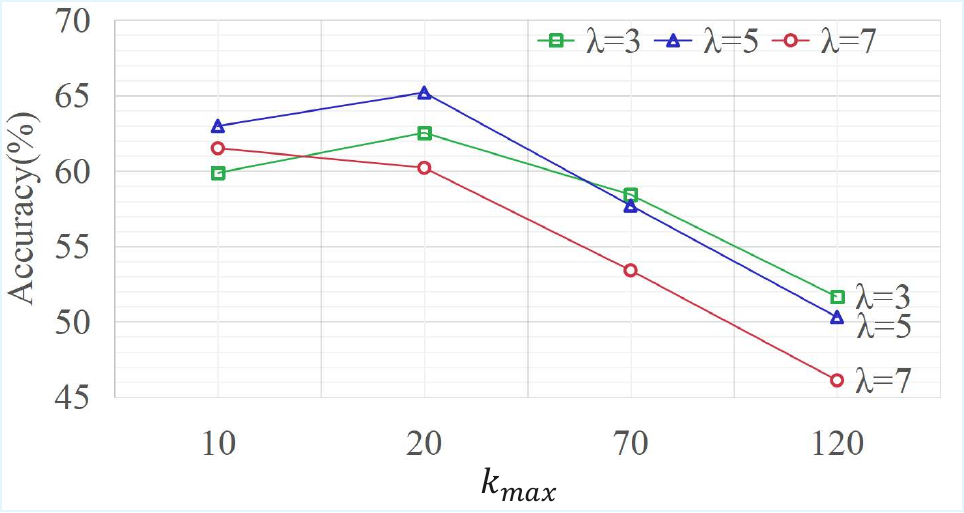}
  \caption{Sensitivity of Accuracy to $k_{\max}$ and $\lambda$.}
  \label{sensitivity}
\end{subfigure}
\caption{Ablation Results on MMRel using LLaVA-1.5: (a) Ablation study; (b) Sensitivity analysis.}
\label{combined}
\end{figure*}

As shown in Table~\ref{ablation}, the full ChainMPQ achieves an accuracy of 65.20\%. Removing text-guided attention reduces accuracy by 1.14\%, suggesting that enhancing subject- and object-related tokens provides limited but measurable benefit in reducing hallucinations. Omitting multi-perspective questions decreases accuracy by 3.68\%, confirming their importance in guiding step-by-step inference, which forms the cornerstone of the chain. Eliminating the interleaved chain lowers accuracy by 3.08\%, indicating that transferring visual memory plays an equally important role in the reasoning process. Despite these degradations, all three ablated variants still outperform the baseline LLaVA (59.02\% accuracy), while the complete ChainMPQ demonstrates the strongest effect.  

\subsubsection{Sensitivity analysis of hyperparameters}
We also analyze the sensitivity of $k_{\text{max}}$ , the maximum value for top-K in Eq.~\ref{topk}. We set $k_{\text{max}}$ to 5\%, 10\%, 30\%, and 50\% of the total number of patches, corresponding to approximately 10, 20, 70, and 120 tokens. We also vary the maximum bias weight parameter $\lambda$, which is the coefficient of $\alpha$ in Eq.~\ref{modify attention}, and test values of 3, 5, and 7. As shown in Figure~\ref{sensitivity}, performance peaks at $\lambda=5$ and $k_{\text{max}}=20$, where the accuracy reaches 65.20\%.When $k_{\text{max}}$ is too large, the model uses almost all image tokens, which weakens its focus on important regions and breaks the normal attention process. In contrast, when $k_{\text{max}}$ is too small, the model may miss key information, especially for questions that rely on scattered visual features. Similarly, if $\lambda$ is too large, the model becomes overly dependent on past memory, which disturbs attention propagation and reduces accuracy. If $\lambda$ is too small, the bias is weak and the performance is close to the original LLaVA baseline.

\subsection{Case Study}

To intuitively illustrate the ChainMPQ process, we present two real examples from MMRel, namely an action case and a spatial case, which together account for over 90\% of the MMRel dataset~\cite{nie2024mmrel}. Due to page limits, another case study of competitive type is provided in Appendix A.6.2.

\begin{figure}[t]
  \centering
  \begin{subfigure}{0.48\textwidth}
    \centering
    \includegraphics[width=\linewidth]{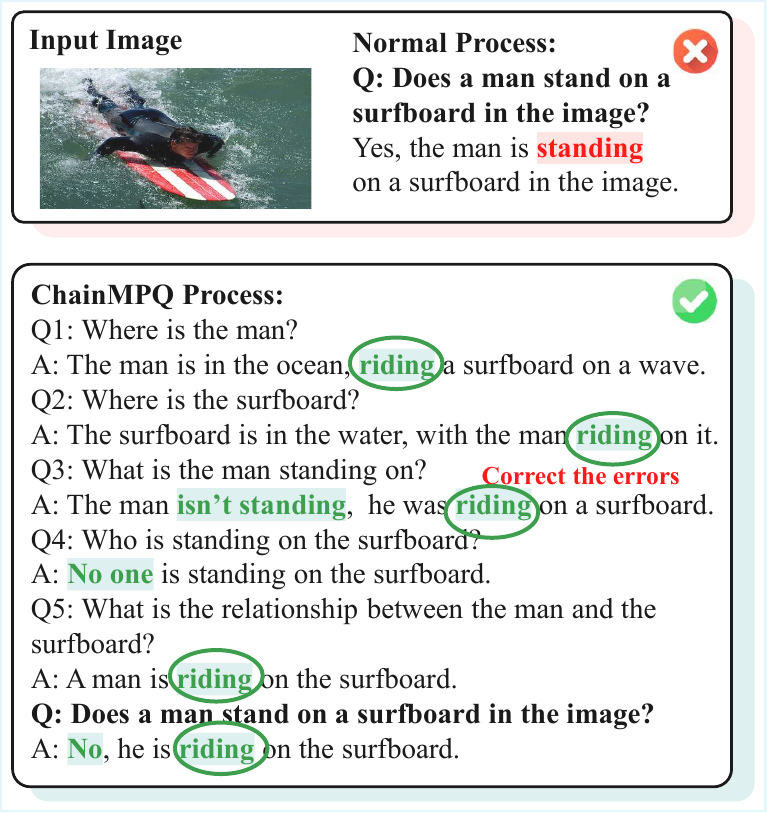}
    \caption{Case study for action category}
    \label{fig:cs_a}
  \end{subfigure}\hfill
  \begin{subfigure}{0.48\textwidth}
    \centering
    \includegraphics[width=\linewidth]{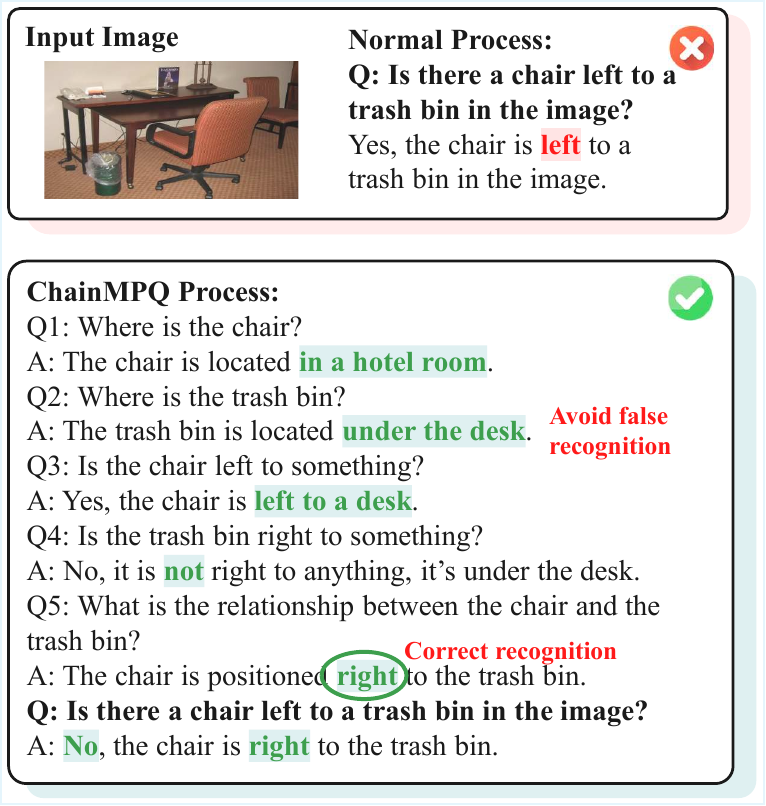}
    \caption{Case study for spatial category}
    \label{fig:cs_b}
  \end{subfigure}
  \caption{The full ChainMPQ answering process vs. directly output}
  \label{case_study_text}
\end{figure}
\begin{figure}[t]
  \centering
  \begin{subfigure}{0.48\textwidth}
    \centering
    \includegraphics[width=\linewidth]{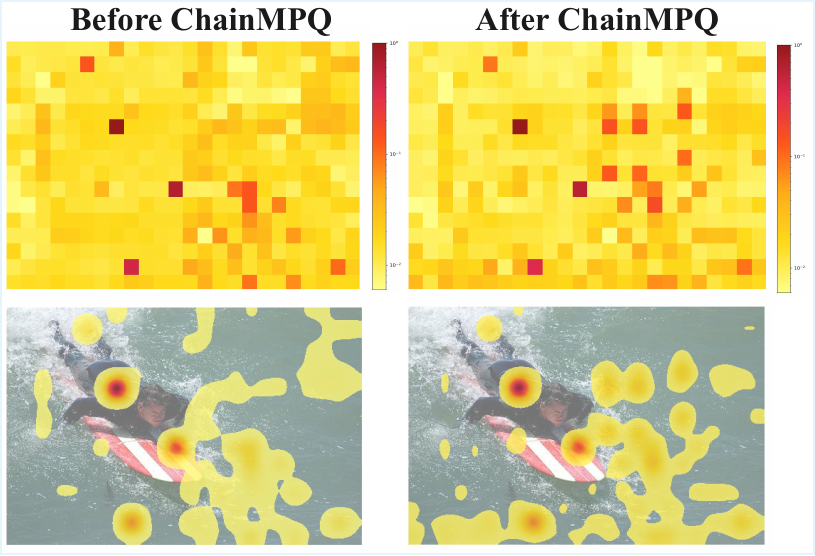}
    \caption{Case study for action category}
    \label{fig:cs_a}
  \end{subfigure}\hfill
  \begin{subfigure}{0.48\textwidth}
    \centering
    \includegraphics[width=\linewidth]{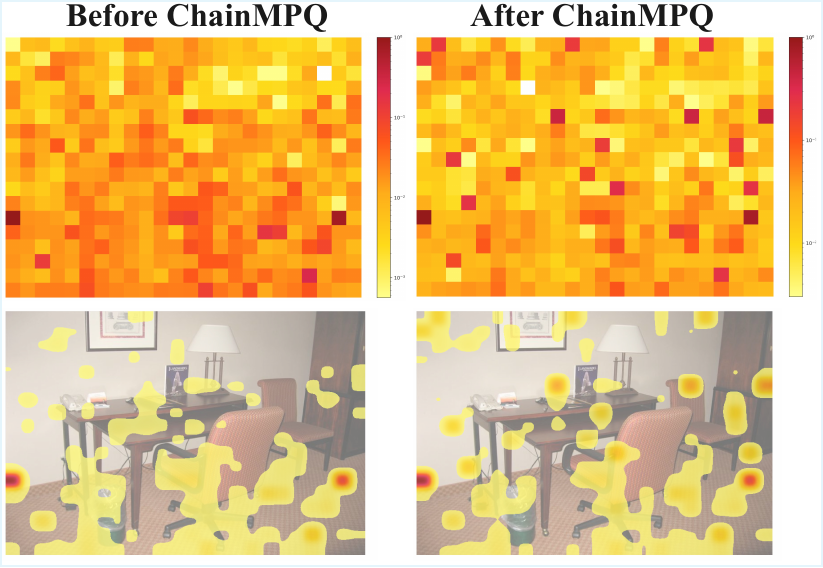}
    \caption{Case study for spatial category}
    \label{fig:cs_b}
  \end{subfigure}
  \caption{Comparison of attention maps between directly answering the original question and answering it using ChainMPQ.}
  \label{case_study_attn}
\end{figure}

Figure~\ref{case_study_text} illustrates the complete processing pipeline of ChainMPQ. In the action case "Does a man stand on a surfboard in the image?", the baseline answers "yes", confusing the true relation "riding" with "standing". ChainMPQ first localizes the subject and the object with two guiding questions, identifies the action “riding,” and propagates this cue through textual context and attention transfer to produce the correct answer "no". In the spatial case "Is there a chair left to a trash bin in the image?", the baseline is again incorrect. ChainMPQ retrieves the locations of the chair and the bin under the desk, constrains subsequent reasoning to the correct regions, and then verifies the directional relation to conclude that the chair is to the "right" of the bin. Across both cases, decomposition helps the model locate the correct regions, reduce false language priors, and follow a clear path to verify relations, leading to answers that aligned with the image.

Figure~\ref{case_study_attn} presents that across both action and spatial cases, ChainMPQ produces attention maps that are more sharply concentrated on task-relevant regions, especially the subject and the object. This pattern indicates that the focused attention triggered by earlier related subquestions is propagated through the chain via textual context and attention transfer, so that when answering the original question the model attends more precisely to the related areas. In both examples, ChainMPQ can suppress attention to irrelevant background areas and focus attention on the interaction between the subject and the target object, aligning the attention trajectory with the queried relation. The resulting focus correlates with corrected predictions and fewer contradictions between the image and the answer, which suggests improved grounding and more faithful relational inference.

\section{CONCLUSION}
We propose \textbf{ChainMPQ} (\textbf{M}ulti-\textbf{P}erspective \textbf{Q}uestions guided Interleaved Text-image Reasoning \textbf{Chain}), a training-free framework that enhances the ability of LVLMs to recognize relationships by utilizing accumulated textual and visual memories. ChainMPQ first strengthens the relevant visual tokens using keywords extracted from the question. It then constructs a set of complementary questions from different perspectives, each closely centered on the subject, the object, and their relationship. These questions are sequentially fed into the model, with both textual and visual memories from earlier rounds propagated to subsequent ones, thereby forming an interleaved chain of image and text. Experimental results demonstrate that ChainMPQ effectively reduces relation hallucinations and improves factuality across multiple LVLMs and benchmarks, providing a simple yet robust framework for step-by-step relational inference.

\section{Future Work}

Although ChainMPQ consistently improves the mitigation of relation hallucinations, it still relies on attention distributions as a proxy for visual evidence, which may not fully capture the underlying reasoning process. Future work will incorporate a causality-based attribution mechanism. By detecting and correcting outputs that contradict the image, it prevents errors from cascading. 

It is worth mentioning that a major challenge in relation hallucination lies in the spatial granularity of visual tokens, which often misalign with real-world object boundaries. This misalignment, observed in lower performance for the spatial category in MMRel in our experiments, can hinder accurate relational reasoning. Future research may alleviate this issue by exploring multi-scale visual representations or by integrating explicit scene graph representations to improve robustness in fine-grained relational understanding.

\section*{Acknowledgments}

The work is partially supported by the U.S. National Science Foundation (NSF) Grant CRII 2451683, an NVIDIA Academic Grants Program, a U.S. Bank Academic Research Award, the University of California, Merced, and a UC Merced Faculty Research Award. 
The views and conclusions are those of the authors and do not necessarily reflect the official policy or position of the U.S. Government.

\bibliography{iclr2026_conference}

@inproceedings{gao2025interleaved,
  title={Interleaved-modal chain-of-thought},
  author={Gao, Jun and Li, Yongqi and Cao, Ziqiang and Li, Wenjie},
  booktitle={Proceedings of the Computer Vision and Pattern Recognition Conference},
  pages={19520--19529},
  year={2025}
}

@article{zheng2024reefknot,
  title={Reefknot: A comprehensive benchmark for relation hallucination evaluation, analysis and mitigation in multimodal large language models},
  author={Zheng, Kening and Chen, Junkai and Yan, Yibo and Zou, Xin and Hu, Xuming},
  journal={arXiv preprint arXiv:2408.09429},
  year={2024}
}

@article{wu2025mitigating,
  title={Mitigating hallucinations in multimodal spatial relations through constraint-aware prompting},
  author={Wu, Jiarui and Liu, Zhuo and He, Hangfeng},
  journal={arXiv preprint arXiv:2502.08317},
  year={2025}
}

@article{bai2024hallucination,
  title={Hallucination of multimodal large language models: A survey},
  author={Bai, Zechen and Wang, Pichao and Xiao, Tianjun and He, Tong and Han, Zongbo and Zhang, Zheng and Shou, Mike Zheng},
  journal={arXiv preprint arXiv:2404.18930},
  year={2024}
}

@inproceedings{yang2025mitigating,
  title={Mitigating hallucinations in large vision-language models via dpo: On-policy data hold the key},
  author={Yang, Zhihe and Luo, Xufang and Han, Dongqi and Xu, Yunjian and Li, Dongsheng},
  booktitle={Proceedings of the Computer Vision and Pattern Recognition Conference},
  pages={10610--10620},
  year={2025}
}

@article{jiang2024interpreting,
  title={Interpreting and editing vision-language representations to mitigate hallucinations},
  author={Jiang, Nick and Kachinthaya, Anish and Petryk, Suzie and Gandelsman, Yossi},
  journal={arXiv preprint arXiv:2410.02762},
  year={2024}
}

@inproceedings{jiang2025devils,
  title={Devils in middle layers of large vision-language models: Interpreting, detecting and mitigating object hallucinations via attention lens},
  author={Jiang, Zhangqi and Chen, Junkai and Zhu, Beier and Luo, Tingjin and Shen, Yankun and Yang, Xu},
  booktitle={Proceedings of the Computer Vision and Pattern Recognition Conference},
  pages={25004--25014},
  year={2025}
}

@article{zhang2025self,
  title={Self-correcting decoding with generative feedback for mitigating hallucinations in large vision-language models},
  author={Zhang, Ce and Wan, Zifu and Kan, Zhehan and Ma, Martin Q and Stepputtis, Simon and Ramanan, Deva and Salakhutdinov, Russ and Morency, Louis-Philippe and Sycara, Katia and Xie, Yaqi},
  journal={arXiv preprint arXiv:2502.06130},
  year={2025}
}

@article{wu2025generate,
  title={Generate, but Verify: Reducing Hallucination in Vision-Language Models with Retrospective Resampling},
  author={Wu, Tsung-Han and Lee, Heekyung and Ge, Jiaxin and Gonzalez, Joseph E and Darrell, Trevor and Chan, David M},
  journal={arXiv preprint arXiv:2504.13169},
  year={2025}
}

@article{wang2024mllm,
  title={Mllm can see? dynamic correction decoding for hallucination mitigation},
  author={Wang, Chenxi and Chen, Xiang and Zhang, Ningyu and Tian, Bozhong and Xu, Haoming and Deng, Shumin and Chen, Huajun},
  journal={arXiv preprint arXiv:2410.11779},
  year={2024}
}

@article{xie2025fg,
  title={FG-CLIP: Fine-Grained Visual and Textual Alignment},
  author={Xie, Chunyu and Wang, Bin and Kong, Fanjing and Li, Jincheng and Liang, Dawei and Zhang, Gengshen and Leng, Dawei and Yin, Yuhui},
  journal={arXiv preprint arXiv:2505.05071},
  year={2025}
}

@article{lin2025mind,
  title={Mind with eyes: from language reasoning to multimodal reasoning},
  author={Lin, Zhiyu and Gao, Yifei and Zhao, Xian and Yang, Yunfan and Sang, Jitao},
  journal={arXiv preprint arXiv:2503.18071},
  year={2025}
}

@article{qin2025uni,
  title={Uni-cot: Towards Unified Chain-of-Thought Reasoning Across Text and Vision},
  author={Qin, Luozheng and Gong, Jia and Sun, Yuqing and Li, Tianjiao and Yang, Mengping and Yang, Xiaomeng and Qu, Chao and Tan, Zhiyu and Li, Hao},
  journal={arXiv preprint arXiv:2508.05606},
  year={2025}
}

@article{wei2022chain,
  title={Chain-of-thought prompting elicits reasoning in large language models},
  author={Wei, Jason and Wang, Xuezhi and Schuurmans, Dale and Bosma, Maarten and Xia, Fei and Chi, Ed and Le, Quoc V and Zhou, Denny and others},
  journal={Advances in neural information processing systems},
  volume={35},
  pages={24824--24837},
  year={2022}
}

@article{nie2024mmrel,
  title={Mmrel: A relation understanding dataset and benchmark in the mllm era},
  author={Nie, Jiahao and Zhang, Gongjie and An, Wenbin and Tan, Yap-Peng and Kot, Alex C and Lu, Shijian},
  journal={arXiv e-prints},
  pages={arXiv--2406},
  year={2024}
}

@misc{bai2025qwen25vltechnicalreport,
      title={Qwen2.5-VL Technical Report}, 
      author={Shuai Bai and Keqin Chen and Xuejing Liu and Jialin Wang and Wenbin Ge and Sibo Song and Kai Dang and Peng Wang and Shijie Wang and Jun Tang and Humen Zhong and Yuanzhi Zhu and Mingkun Yang and Zhaohai Li and Jianqiang Wan and Pengfei Wang and Wei Ding and Zheren Fu and Yiheng Xu and Jiabo Ye and Xi Zhang and Tianbao Xie and Zesen Cheng and Hang Zhang and Zhibo Yang and Haiyang Xu and Junyang Lin},
      year={2025},
      eprint={2502.13923},
      archivePrefix={arXiv},
      primaryClass={cs.CV},
      url={https://arxiv.org/abs/2502.13923}, 
}

@inproceedings{liu2024improved,
  title={Improved baselines with visual instruction tuning},
  author={Liu, Haotian and Li, Chunyuan and Li, Yuheng and Lee, Yong Jae},
  booktitle={Proceedings of the IEEE/CVF conference on computer vision and pattern recognition},
  pages={26296--26306},
  year={2024}
}

@article{wu2024evaluating,
  title={Evaluating and analyzing relationship hallucinations in large vision-language models},
  author={Wu, Mingrui and Ji, Jiayi and Huang, Oucheng and Li, Jiale and Wu, Yuhang and Sun, Xiaoshuai and Ji, Rongrong},
  journal={arXiv preprint arXiv:2406.16449},
  year={2024}
}

@article{dai2023instructblip,
  title={Instructblip: Towards general-purpose vision-language models with instruction tuning},
  author={Dai, Wenliang and Li, Junnan and Li, Dongxu and Tiong, Anthony and Zhao, Junqi and Wang, Weisheng and Li, Boyang and Fung, Pascale N and Hoi, Steven},
  journal={Advances in neural information processing systems},
  volume={36},
  pages={49250--49267},
  year={2023}
}

@article{li2023evaluating,
  title={Evaluating object hallucination in large vision-language models},
  author={Li, Yifan and Du, Yifan and Zhou, Kun and Wang, Jinpeng and Zhao, Wayne Xin and Wen, Ji-Rong},
  journal={arXiv preprint arXiv:2305.10355},
  year={2023}
}

@article{rohrbach2018object,
  title={Object hallucination in image captioning},
  author={Rohrbach, Anna and Hendricks, Lisa Anne and Burns, Kaylee and Darrell, Trevor and Saenko, Kate},
  journal={arXiv preprint arXiv:1809.02156},
  year={2018}
}

@misc{touvron2023llamaopenefficientfoundation,
      title={LLaMA: Open and Efficient Foundation Language Models}, 
      author={Hugo Touvron and Thibaut Lavril and Gautier Izacard and Xavier Martinet and Marie-Anne Lachaux and Timothée Lacroix and Baptiste Rozière and Naman Goyal and Eric Hambro and Faisal Azhar and Aurelien Rodriguez and Armand Joulin and Edouard Grave and Guillaume Lample},
      year={2023},
      eprint={2302.13971},
      archivePrefix={arXiv},
      primaryClass={cs.CL},
      url={https://arxiv.org/abs/2302.13971}, 
}

@misc{bai2023qwentechnicalreport,
      title={Qwen Technical Report}, 
      author={Jinze Bai and Shuai Bai and Yunfei Chu and Zeyu Cui and Kai Dang and Xiaodong Deng and Yang Fan and Wenbin Ge and Yu Han and Fei Huang and Binyuan Hui and Luo Ji and Mei Li and Junyang Lin and Runji Lin and Dayiheng Liu and Gao Liu and Chengqiang Lu and Keming Lu and Jianxin Ma and Rui Men and Xingzhang Ren and Xuancheng Ren and Chuanqi Tan and Sinan Tan and Jianhong Tu and Peng Wang and Shijie Wang and Wei Wang and Shengguang Wu and Benfeng Xu and Jin Xu and An Yang and Hao Yang and Jian Yang and Shusheng Yang and Yang Yao and Bowen Yu and Hongyi Yuan and Zheng Yuan and Jianwei Zhang and Xingxuan Zhang and Yichang Zhang and Zhenru Zhang and Chang Zhou and Jingren Zhou and Xiaohuan Zhou and Tianhang Zhu},
      year={2023},
      eprint={2309.16609},
      archivePrefix={arXiv},
      primaryClass={cs.CL},
      url={https://arxiv.org/abs/2309.16609}, 
}

@misc{touvron2023llama2openfoundation,
      title={Llama 2: Open Foundation and Fine-Tuned Chat Models}, 
      author={Hugo Touvron and Louis Martin and Kevin Stone and Peter Albert and Amjad Almahairi and Yasmine Babaei and Nikolay Bashlykov and Soumya Batra and Prajjwal Bhargava and Shruti Bhosale and Dan Bikel and Lukas Blecher and Cristian Canton Ferrer and Moya Chen and Guillem Cucurull and David Esiobu and Jude Fernandes and Jeremy Fu and Wenyin Fu and Brian Fuller and Cynthia Gao and Vedanuj Goswami and Naman Goyal and Anthony Hartshorn and Saghar Hosseini and Rui Hou and Hakan Inan and Marcin Kardas and Viktor Kerkez and Madian Khabsa and Isabel Kloumann and Artem Korenev and Punit Singh Koura and Marie-Anne Lachaux and Thibaut Lavril and Jenya Lee and Diana Liskovich and Yinghai Lu and Yuning Mao and Xavier Martinet and Todor Mihaylov and Pushkar Mishra and Igor Molybog and Yixin Nie and Andrew Poulton and Jeremy Reizenstein and Rashi Rungta and Kalyan Saladi and Alan Schelten and Ruan Silva and Eric Michael Smith and Ranjan Subramanian and Xiaoqing Ellen Tan and Binh Tang and Ross Taylor and Adina Williams and Jian Xiang Kuan and Puxin Xu and Zheng Yan and Iliyan Zarov and Yuchen Zhang and Angela Fan and Melanie Kambadur and Sharan Narang and Aurelien Rodriguez and Robert Stojnic and Sergey Edunov and Thomas Scialom},
      year={2023},
      eprint={2307.09288},
      archivePrefix={arXiv},
      primaryClass={cs.CL},
      url={https://arxiv.org/abs/2307.09288}, 
}

@misc{ye2024mplugowlmodularizationempowerslarge,
      title={mPLUG-Owl: Modularization Empowers Large Language Models with Multimodality}, 
      author={Qinghao Ye and Haiyang Xu and Guohai Xu and Jiabo Ye and Ming Yan and Yiyang Zhou and Junyang Wang and Anwen Hu and Pengcheng Shi and Yaya Shi and Chenliang Li and Yuanhong Xu and Hehong Chen and Junfeng Tian and Qi Qian and Ji Zhang and Fei Huang and Jingren Zhou},
      year={2024},
      eprint={2304.14178},
      archivePrefix={arXiv},
      primaryClass={cs.CL},
      url={https://arxiv.org/abs/2304.14178}, 
}

@misc{li2023mimicitmultimodalincontextinstruction,
      title={MIMIC-IT: Multi-Modal In-Context Instruction Tuning}, 
      author={Bo Li and Yuanhan Zhang and Liangyu Chen and Jinghao Wang and Fanyi Pu and Jingkang Yang and Chunyuan Li and Ziwei Liu},
      year={2023},
      eprint={2306.05425},
      archivePrefix={arXiv},
      primaryClass={cs.CV},
      url={https://arxiv.org/abs/2306.05425}, 
}

@misc{bai2023qwenvlversatilevisionlanguagemodel,
      title={Qwen-VL: A Versatile Vision-Language Model for Understanding, Localization, Text Reading, and Beyond}, 
      author={Jinze Bai and Shuai Bai and Shusheng Yang and Shijie Wang and Sinan Tan and Peng Wang and Junyang Lin and Chang Zhou and Jingren Zhou},
      year={2023},
      eprint={2308.12966},
      archivePrefix={arXiv},
      primaryClass={cs.CV},
      url={https://arxiv.org/abs/2308.12966}, 
}

@misc{li2023blip2bootstrappinglanguageimagepretraining,
      title={BLIP-2: Bootstrapping Language-Image Pre-training with Frozen Image Encoders and Large Language Models}, 
      author={Junnan Li and Dongxu Li and Silvio Savarese and Steven Hoi},
      year={2023},
      eprint={2301.12597},
      archivePrefix={arXiv},
      primaryClass={cs.CV},
      url={https://arxiv.org/abs/2301.12597}, 
}

@misc{lai2024lisareasoningsegmentationlarge,
      title={LISA: Reasoning Segmentation via Large Language Model}, 
      author={Xin Lai and Zhuotao Tian and Yukang Chen and Yanwei Li and Yuhui Yuan and Shu Liu and Jiaya Jia},
      year={2024},
      eprint={2308.00692},
      archivePrefix={arXiv},
      primaryClass={cs.CV},
      url={https://arxiv.org/abs/2308.00692}, 
}

@misc{laurençon2024buildingbetterunderstandingvisionlanguage,
      title={Building and better understanding vision-language models: insights and future directions}, 
      author={Hugo Laurençon and Andrés Marafioti and Victor Sanh and Léo Tronchon},
      year={2024},
      eprint={2408.12637},
      archivePrefix={arXiv},
      primaryClass={cs.CV},
      url={https://arxiv.org/abs/2408.12637}, 
}

@misc{li2025enhancingvisionlanguagecompositionalunderstanding,
      title={Enhancing Vision-Language Compositional Understanding with Multimodal Synthetic Data}, 
      author={Haoxin Li and Boyang Li},
      year={2025},
      eprint={2503.01167},
      archivePrefix={arXiv},
      primaryClass={cs.CV},
      url={https://arxiv.org/abs/2503.01167}, 
}

@misc{peng2024synthesizediagnoseoptimizefinegrained,
      title={Synthesize, Diagnose, and Optimize: Towards Fine-Grained Vision-Language Understanding}, 
      author={Wujian Peng and Sicheng Xie and Zuyao You and Shiyi Lan and Zuxuan Wu},
      year={2024},
      eprint={2312.00081},
      archivePrefix={arXiv},
      primaryClass={cs.CV},
      url={https://arxiv.org/abs/2312.00081}, 
}

@misc{liu2024surveyhallucinationlargevisionlanguage,
      title={A Survey on Hallucination in Large Vision-Language Models}, 
      author={Hanchao Liu and Wenyuan Xue and Yifei Chen and Dapeng Chen and Xiutian Zhao and Ke Wang and Liping Hou and Rongjun Li and Wei Peng},
      year={2024},
      eprint={2402.00253},
      archivePrefix={arXiv},
      primaryClass={cs.CV},
      url={https://arxiv.org/abs/2402.00253}, 
}

@misc{shahgir2024illusionvqachallengingopticalillusion,
      title={IllusionVQA: A Challenging Optical Illusion Dataset for Vision Language Models}, 
      author={Haz Sameen Shahgir and Khondker Salman Sayeed and Abhik Bhattacharjee and Wasi Uddin Ahmad and Yue Dong and Rifat Shahriyar},
      year={2024},
      eprint={2403.15952},
      archivePrefix={arXiv},
      primaryClass={cs.CV},
      url={https://arxiv.org/abs/2403.15952}, 
}

@misc{wang2025internvl35advancingopensourcemultimodal,
      title={InternVL3.5: Advancing Open-Source Multimodal Models in Versatility, Reasoning, and Efficiency}, 
      author={Weiyun Wang and Zhangwei Gao and Lixin Gu and Hengjun Pu and Long Cui and Xingguang Wei and Zhaoyang Liu and Linglin Jing and Shenglong Ye and Jie Shao and Zhaokai Wang and Zhe Chen and Hongjie Zhang and Ganlin Yang and Haomin Wang and Qi Wei and Jinhui Yin and Wenhao Li and Erfei Cui and Guanzhou Chen and Zichen Ding and Changyao Tian and Zhenyu Wu and Jingjing Xie and Zehao Li and Bowen Yang and Yuchen Duan and Xuehui Wang and Zhi Hou and Haoran Hao and Tianyi Zhang and Songze Li and Xiangyu Zhao and Haodong Duan and Nianchen Deng and Bin Fu and Yinan He and Yi Wang and Conghui He and Botian Shi and Junjun He and Yingtong Xiong and Han Lv and Lijun Wu and Wenqi Shao and Kaipeng Zhang and Huipeng Deng and Biqing Qi and Jiaye Ge and Qipeng Guo and Wenwei Zhang and Songyang Zhang and Maosong Cao and Junyao Lin and Kexian Tang and Jianfei Gao and Haian Huang and Yuzhe Gu and Chengqi Lyu and Huanze Tang and Rui Wang and Haijun Lv and Wanli Ouyang and Limin Wang and Min Dou and Xizhou Zhu and Tong Lu and Dahua Lin and Jifeng Dai and Weijie Su and Bowen Zhou and Kai Chen and Yu Qiao and Wenhai Wang and Gen Luo},
      year={2025},
      eprint={2508.18265},
      archivePrefix={arXiv},
      primaryClass={cs.CV},
      url={https://arxiv.org/abs/2508.18265}, 
}

@misc{wei2023chainofthoughtpromptingelicitsreasoning,
      title={Chain-of-Thought Prompting Elicits Reasoning in Large Language Models}, 
      author={Jason Wei and Xuezhi Wang and Dale Schuurmans and Maarten Bosma and Brian Ichter and Fei Xia and Ed Chi and Quoc Le and Denny Zhou},
      year={2023},
      eprint={2201.11903},
      archivePrefix={arXiv},
      primaryClass={cs.CL},
      url={https://arxiv.org/abs/2201.11903}, 
}

@misc{wu2025unifiedtripletlevelhallucinationevaluation,
      title={Unified Triplet-Level Hallucination Evaluation for Large Vision-Language Models}, 
      author={Junjie Wu and Tsz Ting Chung and Kai Chen and Dit-Yan Yeung},
      year={2025},
      eprint={2410.23114},
      archivePrefix={arXiv},
      primaryClass={cs.CV},
      url={https://arxiv.org/abs/2410.23114}, 
}

@misc{liu2024mmbenchmultimodalmodelallaround,
      title={MMBench: Is Your Multi-modal Model an All-around Player?}, 
      author={Yuan Liu and Haodong Duan and Yuanhan Zhang and Bo Li and Songyang Zhang and Wangbo Zhao and Yike Yuan and Jiaqi Wang and Conghui He and Ziwei Liu and Kai Chen and Dahua Lin},
      year={2024},
      eprint={2307.06281},
      archivePrefix={arXiv},
      primaryClass={cs.CV},
      url={https://arxiv.org/abs/2307.06281}, 
}

@misc{fu2025mmecomprehensiveevaluationbenchmark,
      title={MME: A Comprehensive Evaluation Benchmark for Multimodal Large Language Models}, 
      author={Chaoyou Fu and Peixian Chen and Yunhang Shen and Yulei Qin and Mengdan Zhang and Xu Lin and Jinrui Yang and Xiawu Zheng and Ke Li and Xing Sun and Yunsheng Wu and Rongrong Ji and Caifeng Shan and Ran He},
      year={2025},
      eprint={2306.13394},
      archivePrefix={arXiv},
      primaryClass={cs.CV},
      url={https://arxiv.org/abs/2306.13394}, 
}
\bibliographystyle{iclr2026_conference}

\appendix
\newpage
\section{Appendix}
\subsection{Detailed Algorithm}
Algorithm 2 describes the complete execution process of ChainMPQ in detail.
\begin{algorithm}[h]
\small
\caption{ChainMPQ (training-free, interleaved text--image reasoning)}
\label{alg:chainmpq}
\KwIn{Image $I$, relational question $Q$}
\KwOut{Final answer $A$}
\KwRet{$(\mathrm{Attn}_i,\ M_i)$}
\KwData{Visual tokens $V$, enhanced visual tokens $V'$, complementary questions $Q_{1:5}$, 
        textual memory $\mathcal{T}$ (accumulates $(Q_i,A_i)$), 
        visual memory $\mathcal{V}$ (accumulates $M_i$). }
\KwParams{$\lambda$, \ $k_\text{max}$}

\Stage{STEP I: Text-guided Attention Enhancement}\\
1.\ Extract subject/object keywords $K$ from $Q$.\;
2.\ $V \gets \mathrm{VisualEncoder}(I)$,\quad $X \gets \mathrm{TextEncoder}(K)$.\;
3.\ Use cross attention on $X$ and $V$ to enhance visual tokens .\;

\Stage{STEP II: Multi-Perspective Prompt Construction}\\
4.\ Extract subject $[S]$, object $[O]$, relation $[R]$ from $Q$.\;
5.\ Build five complementary questions $Q_{1:5}$ with mask strategy:\;
\hspace*{1.2em}$Q_1$: Where is [S]? \;
\hspace*{1.2em}$Q_2$: Where is [O]? \;
\hspace*{1.2em}$Q_3$: What is [S] [R]? \ (mask object)\; 
\hspace*{1.2em}$Q_4$: What is [R] [O]? \ (mask subject)\; 
\hspace*{1.2em}$Q_5$: What is the relationship between [S] and [O]? \ (mask relation)\;
6.\ Initialize $\mathcal{T}\gets\varnothing$,  $\mathcal{V}\gets\varnothing$.;

\Stage{STEP III: Interleaved Text--Image Reasoning}\;
7.\ Direct answers for $Q_1,Q_2$ (no context, no bias):\;
\hspace*{1.2em}For $i\in\{1,2\}$:\quad
$A_i \gets Q_i$ using $V'$;\ 
$\mathcal{T} \gets \mathcal{T}\cup\{(Q_i,A_i)\}$\;

8.\ Context- and mask-aware answers for $Q_3\text{--}Q_5$:\;
\hspace*{1.2em}For $i=3..5$:\;
\hspace*{2.4em}(a) Compute aggregated attention $\mathrm{Attn}_i$($\lambda$).\;
\hspace*{2.4em}(b) Form top-$k$ set and build $M_i$($k_\text{max}$):\;
\hspace*{2.4em}(c) Decode with textual \& visual memory:\;
\hspace*{3.6em}$A_i \gets Q_i$ using $V'$, $\mathcal{T}$; $M_i$, 
$\mathcal{T} \gets \mathcal{T}\cup\{(Q_i,A_i)\}$; $\mathcal{V} \gets \mathcal{V}\cup\{M_i\}$.\;

9.\ Final answer: $A \gets Q$,\ $\mathcal{T},\ \mathcal{V}$. \;
\end{algorithm}

\subsection{Dataset Details}
\textbf{MMRel:} A vision-language benchmark specifically designed to assess and improve large multimodal models' understanding of inter-object relationships. approximately 10.14K Yes/No question–answer pairs and covers three relation types: spatial, action, and comparative. MMRel incorporates both real-world dataset(Visual Genome) and synthetic images (via SDXL and DALL·E), allowing for diverse relational reasoning scenarios, including adversarial and counter-intuitive cases. The dataset is constructed through a combination of GPT-4V–assisted annotations and human validation to ensure high-quality labels. Beyond evaluation, MMRel serves as a valuable resource for instruction tuning and fine-tuning, showing clear improvements in relational grounding when used for training. It reveals that many current vision-language models rely heavily on linguistic priors, often hallucinating relations not supported by visual evidence. 

\textbf{R-Bench:} A diagnostic benchmark focused on detecting relationship hallucinations in large vision-language models (LVLMs). It comprises image-level and instance-level yes/no questions that test a model’s ability to identify whether specific relations between objects truly exist. The benchmark categorizes hallucinations into three co-occurrence types: relation–relation, subject–relation, and relation–object to isolate sources of error. R‑Bench samples are sourced primarily from NoCaps, ensuring minimal training overlap. Experiments show that LVLMs frequently over-rely on commonsense priors, misjudging relations in visually ambiguous or adversarial contexts. As a result, R‑Bench is especially effective for fine-grained analysis of relational reasoning errors, offering insights into model biases and areas for improvement. We clarify that our experiments use the image-level setup of R-Bench. The image-level and instance-level setups share almost the same images and questions, but the instance-level setup also provides bounding box coordinates and focuses more on the relationship between specific object instances. Since most datasets do not provide bounding boxes, and locating the objects involved in a relation is itself an essential ability of LVLMs, we choose the image-level setup for our main experiments to give a fair and clear evaluation of how ChainMPQ improves relational understanding.
\subsection{implementation details of multi-perspective question generation}
First, we use an NLP tool to extract the subject \textit{S}, object \textit{O}, and relation \textit{R} from the original question. For example, the question ``Does the dog chase a disc in the image?'' can be decomposed into the subject ``dog'', the object ``disc'', and the relation ``chase''. We then build two questions to locate the subject and the object. In this example, they are ``Q1: Where is a dog in the image?'' and ``Q2: Where is a disc in the image?''

Next, we apply a masking strategy. In each step, one of the three elements is masked, and the other two are used to form a new question. This helps the model focus on the remaining elements and check the original relation from different views. For this example, after masking and reconstruction, we obtain three new questions: ``Q3: What is the dog chasing?'', ``Q4: What is the disc being chased by?'', and ``Q5: What is the relationship between the dog and the disc?'' We use GPT-3.5 Turbo to refine each question so that the wording is clear and grammatically correct.

\subsection{Experiment Implementation Details}
In the \textbf{Construction of Multi-Perspective Aware Text Prompts} step, we employ the spaCy NLP toolkit to extract salient keywords from the input questions. These keywords are then used to refine the generated prompts, ensuring that the decomposed questions remain semantically coherent and closely aligned with the original query.  

In the \textbf{Construction of the Interleaved Chain of Image and Text} step, we set $n$ as 3 to capture the last 3 layers' attention in decoder. We set $k_{\text{max}}$ to 10\% of the total number of visual patches (i.e., $k_{\text{max}}=20$). This proportion provides a balance between capturing sufficient visual context and avoiding noise from irrelevant regions. The maximum bias weight parameter $\lambda$ (the coefficient of $\alpha$) is fixed at 5, which we found to be a stable value across different datasets and models.  


Regarding the comparative experiments with other methods, we  reproduced both baseline methods using the latest public code and their official instructions. We only replaced the base models with LLaVA-1.5-7B and InstructBLIP-7B, and we changed the datasets to MMRel and R-Bench to match our setting. All comparisons are made under a fully fair setup. For Constraint-Aware Prompting~\cite{wu2025mitigating}, we used the combined-prompt, which is Prompt 10, as it gives the best accuracy.
For Detect-then-Calibrate~\cite{zheng2024reefknot}, we ran their released scripts without any change to the parameters.

\subsection{Experiments on general reasoning benchmarks}
To explore whether improved relation reasoning also lead to better general reasoning or broader application outcomes. We also evaluated ChainMPQ on two general multimodal benchmarks, MMBench~\cite{liu2024mmbenchmultimodalmodelallaround} and MME~\cite{fu2025mmecomprehensiveevaluationbenchmark}. And the results show that ChainMPQ brings consistent gains not only in relation reasoning but also in broader multimodal abilities.

These tasks cover a broad range of multimodal abilities, such as visual logic reasoning, scene understanding, OCR, and object grounding, and are not limited to relation reasoning. Specifically, MMBench defines three capability levels and twenty fine-grained ability dimensions for evaluating different reasoning and perception skills of MLLMs. MME includes tests for both perception and cognition. The perception part covers coarse- and fine-grained skills such as object existence, counting, location, color recognition, poster recognition, celebrity recognition, scene and landmark identification, and artwork recognition. The cognition part includes commonsense reasoning, numerical reasoning, translation, and code reasoning. It contains fourteen sub-tasks in total.

The MMBench results are shown in Table~\ref{tab:mmbench}. Here Overall is the weighted average over all sub-tasks and is the most important indicator of general ability. RR (Relation Reasoning) refers to relation reasoning. The MME results are shown in Table~\ref{tab:mme}. Here Overall is the average score across all fourteen tasks. Perception measures detection, counting, color, position, and OCR skills, where we do not expect large gains.Cognition measures high-level semantic understanding such as reasoning, description, and complex question answering.

In sum, these results show that improving relation reasoning does lead to better general reasoning. ChainMPQ brings consistent gains on the Overall scores of both MMBench and MME, especially in the Relation Reasoning and Cognition part, confirming that stronger relation understanding supports broader multimodal reasoning abilities.

\begin{table}[]
\centering
\caption{\textbf{MMBench} results (Overall \& Relation Reasoning). Higher is better.}
\label{tab:mmbench}

\begin{tabular}{l c c c c}
\toprule
Model & Overall $\uparrow$ & $\Delta$Overall & RR $\uparrow$ & $\Delta$RR \\
\midrule
LLaVA-v1.5-7B   & 66.5 & --   & 58.8 & --   \\
+ ChainMPQ      & 67.8 & +1.3 & 61.3 & +2.5 \\
InstructBLIP-7B & 63.9 & --   & 52.5 & --   \\
+ ChainMPQ      & 65.5 & +1.6 & 55.2 & +2.7 \\
Qwen2.5-VL-7B   & 83.2 & --   & 78.2 & --   \\
+ ChainMPQ      & 84.7 & +1.5 & 81.8 & +3.6 \\
InternVL3-8B    & 83.6 & --   & 82.5 & --   \\
+ ChainMPQ      & 84.2 & +0.6 & 83.9 & +1.4 \\
\bottomrule
\end{tabular}

\end{table}
\begin{table}[t]
\centering
\caption{\textbf{MME} results (Overall, Perception, Cognition). Higher is better.}
\label{tab:mme}

\begin{tabular}{l c c c c c c}
\toprule
Model & Overall $\uparrow$ & $\Delta$Overall 
      & Perception $\uparrow$ & $\Delta$Perc 
      & Cognition $\uparrow$ & $\Delta$Cog \\
\midrule
LLaVA-v1.5-7B   & 1808.4 & --    & 1506.2 & --    & 302.1 & --    \\
+ ChainMPQ      & 1898.3 & +89.9 & 1545.6 & +39.4 & 352.7 & +50.6 \\
InstructBLIP-7B & 1391.4 & --    & 1137.1 & --    & 254.3 & --    \\
+ ChainMPQ      & 1484.6 & +87.2 & 1176.0 & +38.9 & 308.6 & +54.3 \\
Qwen2.5-VL-7B   & 2312.1 & --    & 1698.1 & --    & 613.9 & --    \\
+ ChainMPQ      & 2350.8 & +38.7 & 1719.2 & +21.1 & 631.5 & +17.6 \\
InternVL3-8B    & 2422.0 & --    & 1748.4 & --    & 673.6 & --    \\
+ ChainMPQ      & 2457.1 & +35.1 & 1771.4 & +23.0 & 685.7 & +12.1 \\
\bottomrule
\end{tabular}

\end{table}

\subsection{Detailed Case Study}
\subsubsection{Attention Analysis in Case Study}

\begin{figure}[t]
    \centering
    \includegraphics[width=1.0\textwidth]{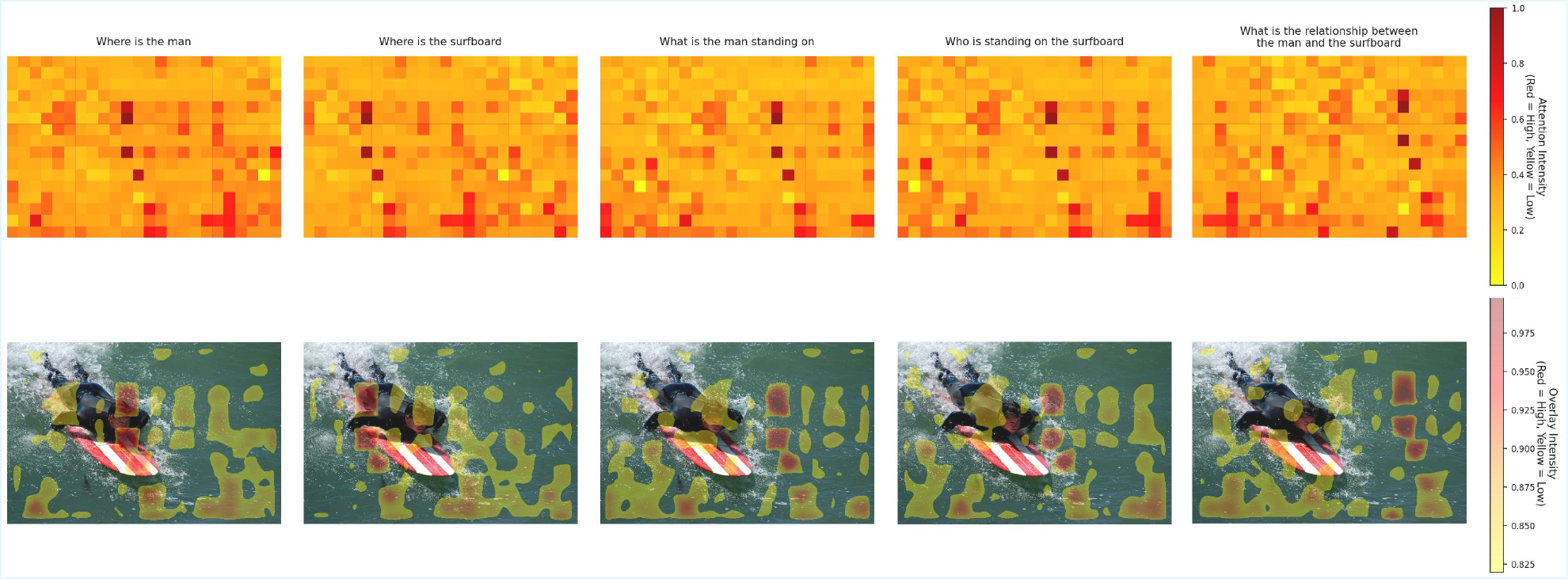}
    \caption{For each generated question, the attention map during the model's answer generation process using ChainMPQ is shown.}
    \label{appen-attn}
\end{figure}

As shown in Figure~\ref{appen-attn}, ChainMPQ progressively guides the model’s reasoning from object localization to relational understanding. The first two sub-questions help the model correctly identify the man and the surfboard, and the attention maps show a clear shift toward the regions that match each answer. The following three sub-questions gradually move the attention toward the interaction between these objects, which helps the model understand the relationship more clearly.

Compared with the direct-answer setting, where the attention maps are scattered and often point to irrelevant areas, ChainMPQ produces more focused and meaningful patterns. In particular, for the question “What is the man standing on”, the attention highlights the surfboard instead of the water. This prevents the model from being confused by the wording of the original question.

This case shows that ChainMPQ not only breaks down a complex question into simple and clear steps but also adjusts the attention process in a helpful way. By guiding the model step by step, ChainMPQ improves both the reasoning path and the final answer, which leads to more stable and accurate relational predictions.

\subsubsection{Case Study for competitive category in MMRel}
 
\begin{figure}[t]
    \centering
    \includegraphics[width=0.7\textwidth]{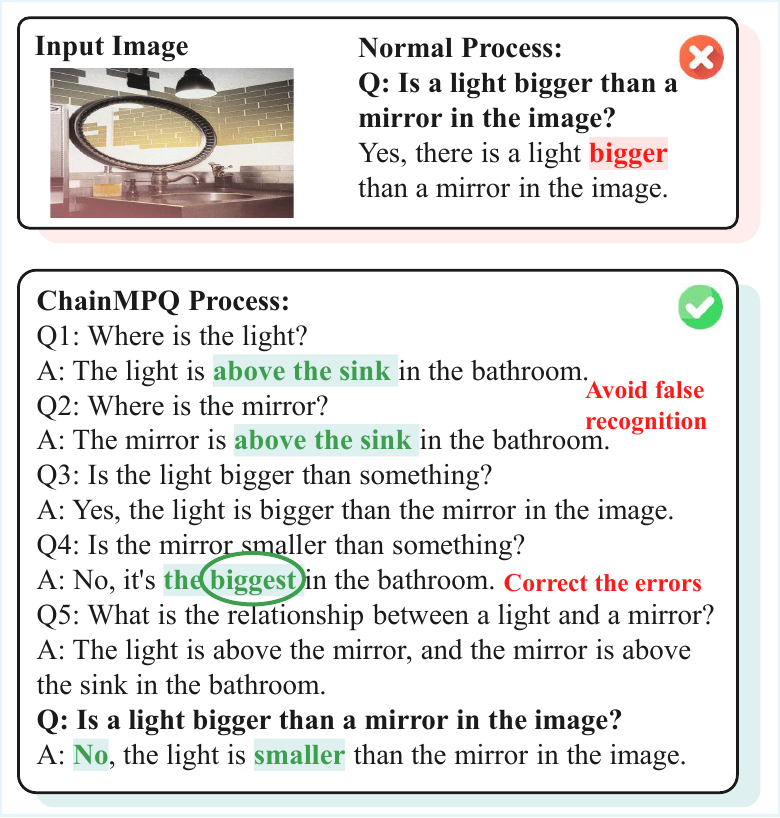}
    \caption{Case study for competitive category}
    \label{competitive_casestudy}
\end{figure}

 We only showed two case studies in the main text because of space limits. This does not mean that our method works better on action or spatial questions than on other types. ChainMPQ is also effective on competitive cases. In Figure~\ref{competitive_casestudy}, we present an example from the competitive domain. This case shows that ChainMPQ can handle competitive questions with the same level of stability as action and spatial questions. It also suggests that our method is not restricted to one type of relation and can work well across different scenarios.

\end{document}